\documentclass[10pt]{article} %

\usepackage[accepted]{rlj} %

\usepackage{amssymb}            %
\usepackage{mathtools}          %
\usepackage{mathrsfs}           %
\usepackage{graphicx}           %
\usepackage{subcaption}         %
\usepackage[space]{grffile}     %
\usepackage{url}                %
\usepackage{lipsum}             %

\usepackage[colorinlistoftodos]{todonotes}

\usepackage{wrapfig}

\usepackage{booktabs}

\definecolor{softbluegray}{HTML}{F2F6F9}
\definecolor{softbluegray1}{HTML}{D9E2EC}
\definecolor{softbluegray2}{HTML}{BCCCDC}
\definecolor{softbluegray3}{HTML}{607992}

\title{Remembering the Markov Property in Cooperative MARL}

\setrunningtitle{Remembering the Markov property in Cooperative MARL}

\author{Kale-ab Abebe Tessera\textsuperscript{1}, Leonard Hinckeldey\textsuperscript{1}, Riccardo Zamboni\textsuperscript{2}, \\ David Abel\textsuperscript{1}, Amos Storkey\textsuperscript{1}}

\emails{\{k.tessera,l.hinckeldey,david.abel,a.storkey\}@ed.ac.uk,\\riccardo.zamboni@polimi.it}

\affiliations{
$^{1}$\textbf{University of Edinburgh, Edinburgh, UK}\\
$^{2}$\textbf{Politecnico di Milano, Milano, Italy}\\
\par %
}

\contribution{
    Provide a succinct but precise list of the contribution(s) of the paper. Use contextual notes to avoid implications of contributions more significant than intended and to clarify and situate the contribution relative to prior work (see the examples below). If there is no additional context, enter ``None''. Try to keep each contribution to a single sentence, although multiple sentences are allowed when necessary. If using complete sentences, include punctuation. If using a single sentence fragment, you may omit the concluding period. A single contribution can be sufficient, and there is no limit on the number of contributions. Submissions will be judged mostly on the contributions claimed on their cover pages and the evidence provided to support them. Major contributions should not be claimed in the main text if they do not appear on the cover page. Overclaiming can lead to a submission being rejected, so it is important to have well-scoped contribution statements on the cover page.
    }
    {
    None
    }

\contribution{
    The submission template for submissions to RLJ/RLC 2025
    }
    {
    Built from previous RLC/RLJ, ICLR, and TMLR submission templates
    }

\contribution{
    \textit{{[}Example of one contribution and corresponding contextual note for the paper ``Policy gradient methods for reinforcement learning with function approximation'' \citep{Sutton2000}.]}\\ This paper presents an expression for the policy gradient when using function approximation to represent the action-value function.
    }
    {
    Prior work established expressions for the policy gradient without function approximation \citep{Williams1992}.
    }


\begin{document}

\maketitle  %

\begin{abstract}

Cooperative multi-agent reinforcement learning (MARL) is typically formalised as a Decentralised Partially Observable Markov Decision Process (Dec-POMDP), where agents must reason about the environment and other agents' behaviour. In practice, current model-free MARL algorithms use simple recurrent function approximators to address the challenge of reasoning about others using partial information. In this position paper, we argue that the empirical success of these methods is not due to effective Markov signal recovery, but rather to learning simple conventions that bypass environment observations and memory. Through a targeted case study, we show that co-adapting agents can learn brittle conventions, which then fail when partnered with non-adaptive agents. Crucially, the same models can learn grounded policies when the task design necessitates it, revealing that the issue is not a fundamental limitation of the learning models but a failure of the benchmark design. Our analysis also suggests that modern MARL environments may not adequately test the core assumptions of Dec-POMDPs. We therefore advocate for new cooperative environments built upon two core principles: (1) behaviours grounded in observations and (2) memory-based reasoning about other agents, ensuring success requires genuine skill rather than fragile, co-adapted agreements.

\end{abstract}

\section{Introduction}
\label{sec:intro}

In many real-world scenarios, teams of agents must make decisions and cooperate under uncertainty without accessing the full conditions of the environment they act upon. This is remarkably the central challenge of multi-agent learning: contending with imperfect information. Decentralised Partially Observable Markov decision processes~\citep[Dec-POMDPs,][]{bernstein2002complexity,oliehoek2016concise} are the prominent decision-making model in these scenarios, as each agent perceives only their own actions and observations. 

\begin{figure}[tb]
    \centering
    \begin{subfigure}[b]{0.48\textwidth}
        \centering
        \includegraphics[width=\textwidth]{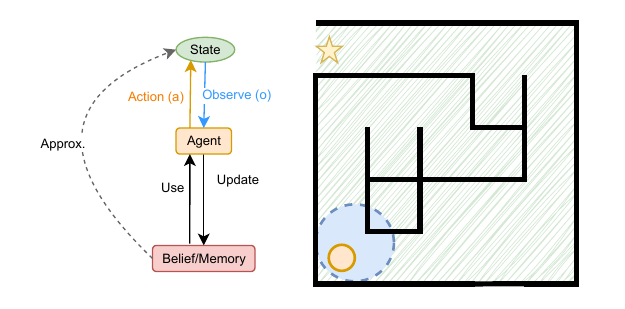}
        \caption{\textbf{POMDP:} \emph{If you can't see, you must remember~\citep{kaelbling1998planning}.}}
        \label{fig:sub1}
    \end{subfigure}%
    \hfill
    \begin{subfigure}[b]{0.48\textwidth}
        \centering
        \includegraphics[width=\textwidth]{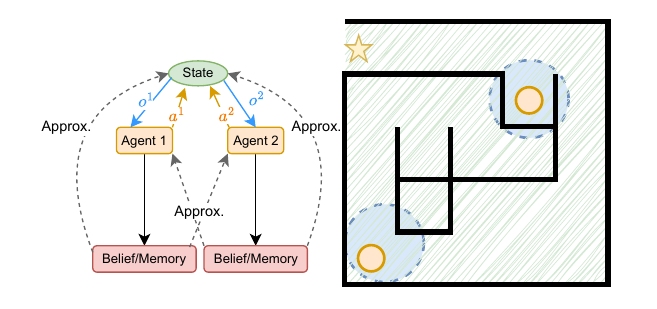}
        \caption{\textbf{Dec-POMDP:} \emph{If you can't see, you must predict~\citep{oliehoek2016concise}}.}
        \label{fig:sub2}
    \end{subfigure}
    \caption{\textbf{Hidden State Requirements in (Dec-)POMDPs.}
    \textbf{(a)} In POMDPs, the agent uses memory or beliefs to approximate the state. 
    \textbf{(b)} In Dec-POMDPs, each agent must additionally predict the behaviour of other agents under uncertainty.}
    \label{fig:subfigureExample}
\end{figure}

As such, they are the more common generalisation of single-agent Partially Observable Markov Decision Processes~\citep[POMDP,][]{aastrom1965optimal, kaelbling1998planning}. Yet, despite their ubiquity in practice, our understanding of MARL~\citep{albrech2024multiagent} in such settings remains limited. This is somewhat expected since even in single-agent settings, planning and learning under partial observability suffer from well-known computational and statistical hardness results~\citep{papadimitriou1987complexity, lusena2001nonapproximability}, and the presence of multiple agents hinders the ability to build a full history of the performed actions and perceived observations, or to recover a distribution over the latent state of the environment, known as \emph{belief}~\citep{kaelbling1998planning}. In principle, each agent should build and update a belief over the joint state and other agents' policies (or equivalently individual histories) -- “multi-agent belief” -- to recover a \textbf{Markovian signal}~\citep{oliehoek2016concise}. As visually represented in Fig.~\ref{fig:subfigureExample}, agents should \textcolor{softbluegray3}{approximate the environment state} and be able to \textcolor{softbluegray3}{predict the behaviour of other agents} to act optimally.

However, exact multi-agent belief-state computation in Dec-POMDPs is known to be NEXP-complete~\citep{bernstein2002complexity}. %
As a result, practical model-free MARL methods rely on finite-memory or recurrent policy representations (e.g., RNNs or GRUs), originally used to handle partial observability in single-agent RL~\citep{hausknecht2015deep}. These are usually instantiated in the \emph{centralised training with decentralised execution} (CTDE)~\citep{oliehoek2008optimal,kraemer2016multi} paradigm, where agents have access to additional information (sometimes from other agents) during training, to improve efficiency.

While the practical success of model-free MARL methods~\citep[among others,][]{yu2022surprising,papoudakis2020benchmarking} might suggest that these methods are adequately recovering the Markov signal by reasoning about the environment and other agents. In this paper, we argue to the contrary. Through a focused case study (Section~\ref{sec:case_study}), we demonstrate that co-adapting agents often sidestep true state recovery by converging on brittle conventions that depend neither on grounded observations nor on the recurrent hidden state. Yet, the same architectures can learn state-grounded policies once the task is not amenable to conventions, hinting that the shortfall lies not in the learning or modelling capacity of the models but in the environment design.

Additionally, we highlight this misalignment in modern benchmarks such as Hanabi~\citep{bard2020hanabi}, MaBrax~\citep{rutherford2023jaxmarl,peng2021facmac} and SMAX~\citep{rutherford2023jaxmarl}, where either memoryless feed-forward policies can paradoxically outperform or match their recurrent counterparts or the learned policies do not adequately require dependence on observations or history (Section~\ref{sec:reasoning}). Such results imply that many MARL tasks do not in fact demand the kind of temporal reasoning and belief maintenance that Dec-POMDP theory considers essential.

Collectively, these findings expose a gap between the reasoning abilities agents possess and the behaviours that current environments test for. Whereas prior work has proposed algorithms to mitigate conventions~\citep{hu2020other,foerster2019bayesian,hu2021off}, we reframe the emergence of conventions as a diagnostic signal: when a benchmark allows effortless coordination through non-generalisable shortcuts, it is the benchmark, rather than the algorithm, that requires re-evaluation. We therefore advocate for new cooperative environments built upon two core principles: \textbf{(1) behaviours grounded in observations} and \textbf{(2) memory-based reasoning} about other agents, ensuring agents must recover and exploit the true Markov structure of MARL problems to succeed.

\section{Related Work}

\paragraph{Conventions.} Prior work has shown that co-trained MARL agents can form conventions that can be brittle to new unseen partners and propose augmenting learning algorithms to tackle this~\citep{hu2020other,foerster2019bayesian,hu2021off}. In contrast, we reframe conventions as a diagnostic signal that the benchmark itself might be ill-posed and fail to test for the intended Dec-POMDP reasoning. Using mutual information metrics, we show that agents can learn to ignore their observations entirely, yet these same methods are capable of learning reliable, grounded policies when the task design necessitates it. Furthermore, although we show some cases of zero-shot coordination failures in Section ~\ref{sec:case_study}, our primary focus (Section ~\ref{sec:reasoning}) is the standard MARL setting where agents are trained and evaluated together.

\paragraph{Environments.} SMAC V2~\citep{ellis2023smacv2} showed that many original SMAC~\citep{samvelyan2019starcraft} maps could be solved by open‑loop policies that ignored local observations (only conditioned on time steps) and introduced "meaningful partial observability" to mitigate this flaw. We show that other modern MARL environments such as MaBrax~\citep{rutherford2023jaxmarl} suffer from an even stronger variant of the same pathology: "blind" agents that receive \emph{no} observations (even without time steps) still obtain non‑trivial returns on a range of configurations (Fig.~\ref{fig:blind_mabrax_detailed} in the Appendix). We extend this line of inquiry by analysing a broader range of environments and looking beyond just partial observability to evaluate the need for grounded, memory-based policies.

\paragraph{Agent Modelling.} Agent modelling techniques aim to predict other agents’ actions, goals, or beliefs to improve coordination or competition~\citep{albrecht2018autonomous}. Modern methods focus on learning latent representations from observation histories, often using auto-encoders with auxiliary prediction losses, to modelling the behaviour of other agents~\citep{papoudakis2021agent,zintgraf2021deep,rabinowitz2018machine,xie2021learning}. While such architectures are powerful, model‑free recurrent policies remain the dominant baseline in cooperative MARL~\citep{yu2022surprising,papoudakis2020benchmarking}. Recent partner modelling work showed that model-free RNNs can encode teammates’ abilities when the environment enables influence over them, in two-player Overcooked environments ~\citep{mon2025partner}. They focused on the ad-hoc teamwork settings with a single controllable agents. Our investigation complements their results as we show memory-based reasoning and grounded policies can emerge when the environment requires it; however, we focus on MARL settings where all agents are co-trained together.

\section{Background}
\label{sec:background}

We next introduce the main concepts that will be covered throughout the paper.

\paragraph*{Interaction Protocol.}~~As a base model for interaction, we consider a discounted Dec-POMDP~\citep{bernstein2002complexity}, defined by the tuple $\mathcal{M} = (\mathcal N, \mathcal{S}, \mathbb{T}, \mathbb{O}, \mu, \{ \mathcal{A}^i \}_{i\in \mathcal N}, \{ \mathcal{O}^i \}_{i \in \mathcal N}, R, \gamma)$. Here, $\mathcal{N}$ is the set of $N \in \mathbb{N}$ agents and $\mathcal{S}$ is the set of global states. At each time step $t$, the system is in some state $s_t \in \mathcal{S}$. Each agent $i \in \mathcal{N}$ selects an action $a_t^i \in \mathcal{A}^i$, forming a joint action $\mathbf{a}_t = (a_t^1, \dots, a_t^N)$ in the joint action space $\mathcal{A} = \prod_{i=1}^N \mathcal{A}^i$. This action leads to a state transition according to the probability function $\mathbb{T}(s_{t+1}|s_t, \mathbf{a}_t)$ and a shared reward $R(s_t, \mathbf{a}_t)$. Agents do not observe the global state $s_t$, instead they receive a local observation $o_t^i \in \mathcal{O}^i$. The joint observation $\mathbf{o}_t$ is drawn according to the observation function $\mathbb{O}(\mathbf{o}_t|s_t, \mathbf{a}_{t-1})$. The goal is to learn a joint policy $\boldsymbol{\pi}$ that maximises the expected discounted return, given an initial state distribution $\mu \in \Delta(\mathcal{S})$ and a discount factor $\gamma \in [0, 1)$:
$
\boldsymbol{\pi}^* = \arg\max_{\boldsymbol{\pi}} \mathbb{E}_{s_0 \sim \mu, \, \mathbf{a}_t \sim \boldsymbol{\pi}} \left[ \sum_{t=0}^{\infty} \gamma^t R(s_t, \mathbf{a}_t) \right].
$

\paragraph*{Markov Property.}~~A Dec-POMDP has the Markov property if the current state contains all relevant information for predicting the future~\citep{sutton1998introduction,oliehoek2016concise}. Formally, this means the transition dynamics do not depend on the full history, $
\mathbb{T}(s_{t+1} \mid s_t, \mathbf{a}_t) = \mathbb{T}(s_{t+1} \mid s_t, \mathbf{a}_t, s_{t-1}, \mathbf{a}_{t-1}, \ldots, s_0, \mathbf{a}_0)$, where $s_t$ is the environment state and $\mathbf{a}_t$ is the joint action. While the dynamics are Markovian in the joint state, individual agents cannot observe this state directly and must act based on partial observations, typically attempting to recover a Markovian signal by forming beliefs or approximations over the state and other agents’ actions~\citep{oliehoek2016concise}.

\paragraph*{Mutual Information.}~~To study the information embedded in our agent's policies, we propose metrics based on mutual information (MI) $\mathbb{I}(X; Y)$, measuring the information shared between two discrete random variables $X$ and $Y$, defined as 
\( \mathbb{I}(X; Y) =  \sum_{X,Y} p(x,y) \log \frac{p(x,y)}{p(x)p(y)}.\)
We measure the MI between an agent's observation and action, $\mathbb{I}(O;A)$, and its recurrent hidden state and action, $\mathbb{I}(H;A)$, to quantify dependence on sensory input and memory, respectively. We estimate these using the k-NN estimator~\citep{kraskov2004estimating,ross2014mutual,garcin2025studying}, while averaging all values across agents.

\section{On the Recovery of Markovian Information in MARL}
\label{sec:case_study}

A plethora of recent works~\citep[among others,][]{yu2022surprising,papoudakis2020benchmarking} provided extensive evidence of the empirical effectiveness of MARL in addressing a wide range of tasks. Such a success story might implicitly suggest that these algorithms do indeed manage to recover essential information for decision-making in multi-agent environments, namely a Markov signal reconstructing the environment's state and other agent's actions. In this section, we address the following fundamental question:

\begin{tcolorbox}[colback=softbluegray, colframe=softbluegray,  boxrule=0.5pt, arc=4pt, width=\linewidth]
\begin{center}
\emph{Are deep MARL policies truly recovering a Markov state? If no, what allows for their success?}
\end{center}
\end{tcolorbox}

In the following, we will control for the information agents have access to so as to analyse the interplay between their ability to recover or extract essential features of the environment and the other agents as well as their ability to solve the tasks. \citet{wijmansemergence,mon2025partner} provided extensive evidence that purely egocentric information, i.e., changes in one's location or orientation, can still give rise to emergent goal-navigation and implicit partner-modelling. We address push this idea further: We construct \emph{blind} environment instantiations, in which agents have access to \emph{no} observation at all. In other words, these blind agents receive no sensory input about the environment, their teammates or even their past actions, and thus they have to rely solely on reward feedback to learn. Concurrently, we will control for their ability to recover a Markovian signal by analysing two different instantiations of Independent PPO~\citep[IPPO,][]{de2020independent}: \textbf{\textcolor{softbluegray3}{MLP (feed-forward)}}, a standard multilayer perceptron that processes each observation independently, and \textbf{\textcolor{softbluegray3}{GRU (recurrent)}}, a gated recurrent unit encoder capable of integrating information over time to form an implicit history representation. All methods do not use parameter sharing. \vspace{-0.3cm}

\paragraph{Warm-up Environment: Prediction Game.}

\begin{figure*}[t]
    \centering
    \begin{subfigure}[b]{0.58\textwidth}
        \centering
        \includegraphics[width=\textwidth]{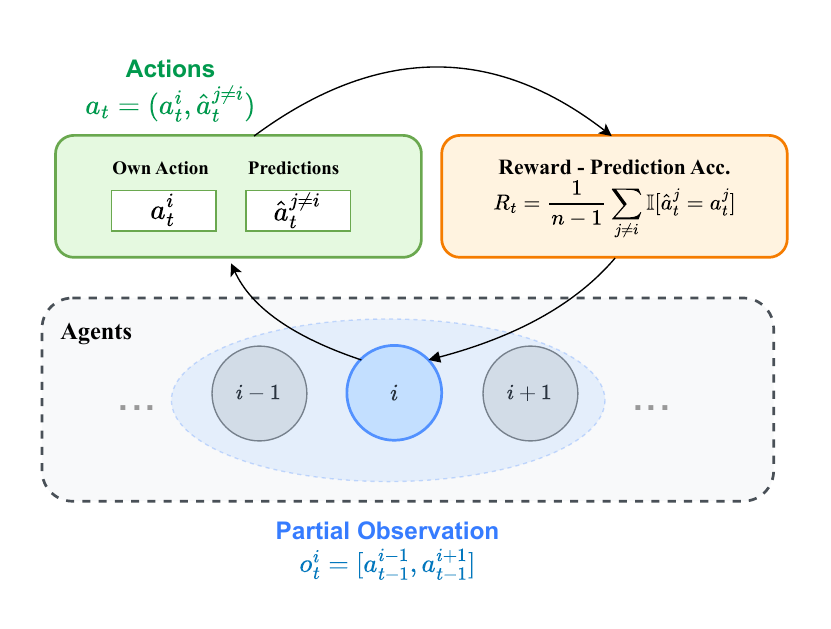}
        \caption{The general agent interaction model, showing the observation-action-reward loop for a single agent within its local neighborhood.}
        \label{fig:main_env_diagram}
    \end{subfigure}%
    \hfill %
    \begin{minipage}[b]{0.38\textwidth}
        \centering
        \vspace{-0.5cm}
        \begin{subfigure}[b]{\linewidth}
            \centering
            \includegraphics[width=\textwidth]{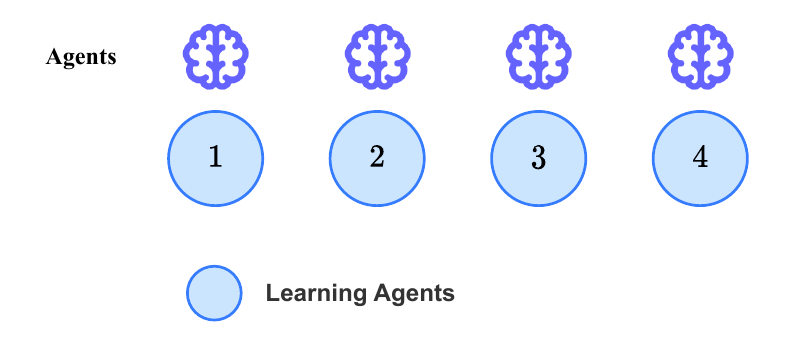}
            \caption{A homogeneous setup with four learning agents.}
            \label{fig:scenario_1}
        \end{subfigure}

        \begin{subfigure}[b]{\linewidth}
            \centering
            \includegraphics[width=\textwidth]{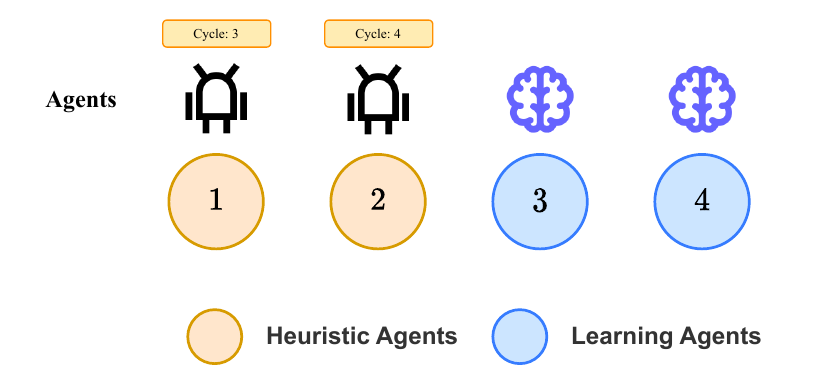}
            \caption{A heterogeneous setup with two learning and two heuristic agents.}
            \label{fig:scenario_2}
        \end{subfigure}
    \end{minipage}
    
    \caption{Overview of Predictive Game environment and the specific agent configurations used in the experiments.}
    \label{fig:env_overview_combined}
\end{figure*}

As a first experiment instantiation, we design the \textbf{\textcolor{softbluegray3}{Prediction Game}} (see Fig.~\ref{fig:main_env_diagram}): A cooperative task involving agents that act in an environment where at each timestep $t$, an agent $i$ receives a partial observation $o_{t}^i$ consisting of the previous actions of its two immediate neighbours, namely $a_{t-1}^{i-1}$ and $a_{t-1}^{i+1}$. Esch agent then selects a multi-discrete action $a^i_t = (\tilde a_t^i, \hat{a}_t^{-i})$, made of two components: \textcolor{softbluegray3}{own action} ($\tilde a_t^i$) being the agent's actual action; \textcolor{softbluegray3}{action prediction} ($\hat{a}_t^{-i}$) being a vector of predictions for the actions of the set of other agents $-i = \{j \in \mathcal N \neq i\}$. Finally, agents share the same reward $R_t$ explicitly defined as its prediction accuracy at timestep $t$, in other words
\(
R_{t} = \frac{1}{N-1}\sum_{i\in \mathcal N} \mathbb{I}[\hat{a}_{t}^{-i} = \tilde a_{t}^{-i}],
\)
where $\mathbb{I}$ is the indicator function. By means of this reward function, Prediction Game ensures that agents will have to accurately model and predict the behaviour of others in order to succeed. In this way, the problem explicitly models Dec-POMDPs instances where recovering the Markovian signal requires agents to not only estimate the environment's state but also to predict the behaviour of others.\vspace{-0.3cm}

\begin{figure}[t]
    \centering

    \begin{subfigure}[t]{0.32\textwidth}
        \centering
        \includegraphics[width=\linewidth]{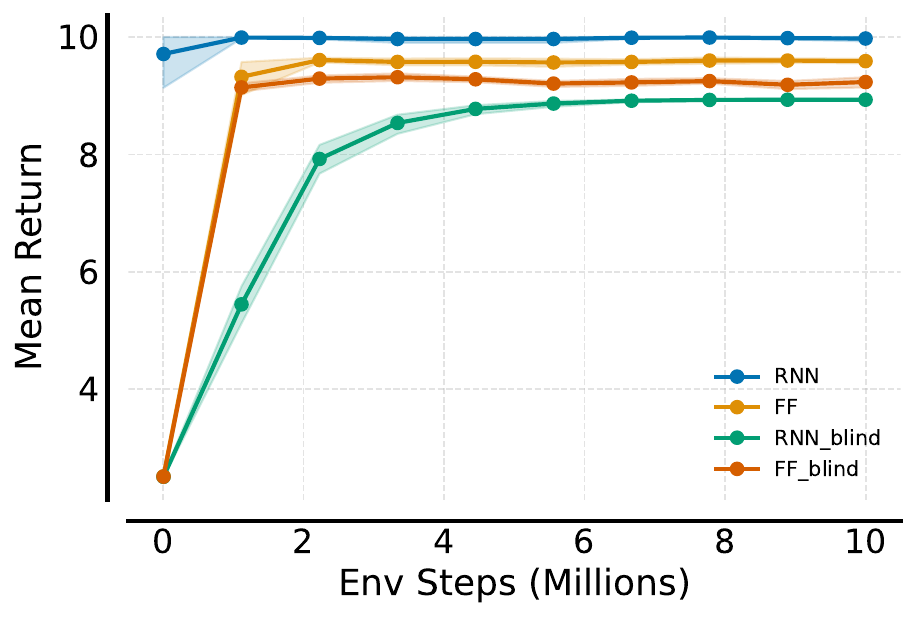}
        \caption{Mean Return, $95\%$ CI.}
        \label{fig:four_return}
    \end{subfigure}
    \hfill
    \begin{subfigure}[t]{0.32\textwidth}
        \centering
        \includegraphics[width=\linewidth]{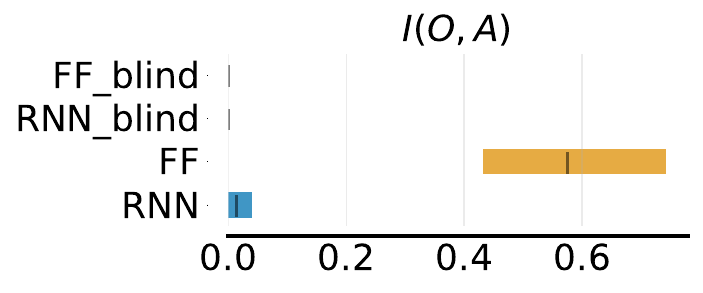}
        \caption{\(I(O, A)\)}
        \label{fig:four_mi_obs_actions}
    \end{subfigure}
    \hfill
    \begin{subfigure}[t]{0.32\textwidth}
        \centering
        \includegraphics[width=\linewidth]{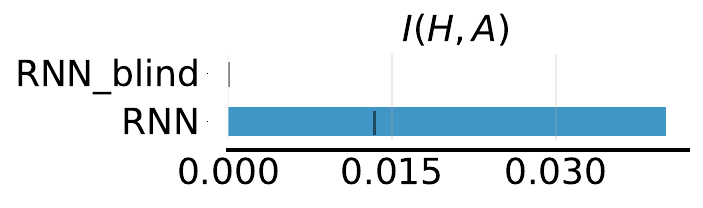}
        \caption{\(I(H, A)\)}
        \label{fig:four_mi_hidden_actions}
    \end{subfigure}

    \caption{Concurrent Learning Agents Experiment: (a) Learning performances %
    ; (b) Mutual Information between actions and observations; (c) Mutual Information between agents' actions and histories. We report the mean and bootstrapped $95\%$ Confidence Intervals (CI) over 10 seeds.}
    \label{fig:four_learning}
\end{figure}

\paragraph*{Emergence of Conventions in Concurrently Learning Agents.}~~First, we addressed homogeneous settings where four IPPO agents learn concurrently (see Fig.~\ref{fig:scenario_1}). As shown in Fig.~\ref{fig:four_return}, learning both recurrent (\textbf{\textcolor{softbluegray3}{RNN}}) and feed-forward (\textbf{\textcolor{softbluegray3}{FF}}) policies lead to near-optimal performances. Strikingly, the \emph{blind} agents, i.e., the ones lacking both memory and observations, also converge to high-performing policies. We claim that this surprising success does not stem from sophisticated, grounded reasoning, but rather agents learn to ignore their observations and form simple, synchronised policies. Indeed, Figures~\ref{fig:four_mi_obs_actions},~\ref{fig:four_mi_hidden_actions} show that the MI between observations and actions $I(O,A)$ and hidden state of the recurrent network $H$ and actions $I(H,A)$ is low, as the maximum MI possible for these experiments is $8$. In other words, the agents are not learning to rely on their grounded observations or hidden state, but rather simple, emergent conventions that bypass the need for complex reasoning (we refer to Figure~\ref {fig:action_dist_pred_game} in Appendix for action distributions plots).\vspace{-0.3cm}

\paragraph*{Failure of Conventions in Concurrently Learning Agents.}~~To test the robustness of the conventions emerging in the previous case, we test resulting policies by substituting two of the agents with fixed-policy agents (we report details on such fixed policies in Appendix). Indeed, this zero-shot coordination task results in breaking the learned conventions and requires agents to infer the partners' behaviour from observation or history alone: The results in Table~\ref{tab:add_heuristic_agents} show a drastic drop in performance for all agent types. This demonstrates that the learned policies are indeed brittle and fail to generalise. The conventions learned during concurrent learning are a shortcut that completely bypasses the challenge of robustly modelling other agents, failing immediately when partners behave unexpectedly.\vspace{-0.3cm}

\begin{table}[t]
\small
\centering
\caption{Performance Comparison: Baseline vs Adding Heuristic Agents}
\label{tab:heuristic_evaluation}
\begin{tabular}{lcccc}
\toprule
\textbf{Scenario} & \textbf{RNN} & \textbf{FF} & \textbf{RNN\_blind} & \textbf{FF\_blind} \\
\midrule
Baseline & 9.97 (9.92, 10.03) & 9.59 (9.55, 9.63) & 8.94 (8.91, 8.97) & 9.23 (9.12, 9.34) \\
Add Heuristic & 4.80 (4.41, 5.20) & 4.04 (3.26, 4.81) & 4.57 (4.18, 4.95) & 4.62 (4.25, 4.99) \\
\bottomrule
\end{tabular}
\label{tab:add_heuristic_agents}
\end{table}

\begin{figure}[t!]
    \centering

    \begin{subfigure}[t]{0.32\textwidth}
        \centering
        \includegraphics[width=\linewidth]{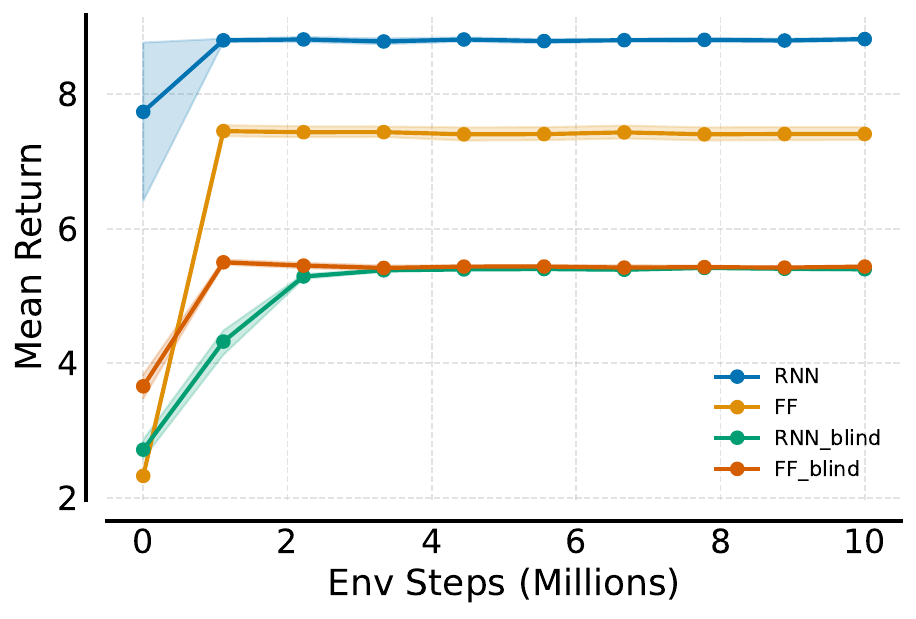}
        \caption{Mean Return, $95\%$ CI. }
        \label{fig:four_learning_agents_sub}
    \end{subfigure}
    \hfill
    \begin{subfigure}[t]{0.32\textwidth}
        \centering
        \includegraphics[width=\linewidth]{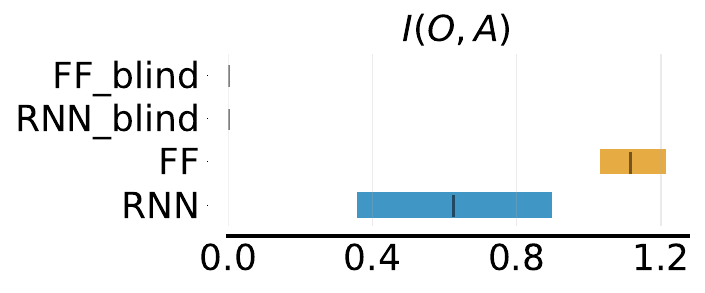}
        \caption{\(I(O, A)\)}
        \label{fig:two_heuristic_agents_sub}
    \end{subfigure}
    \hfill
    \begin{subfigure}[t]{0.32\textwidth}
        \centering
        \includegraphics[width=\linewidth]{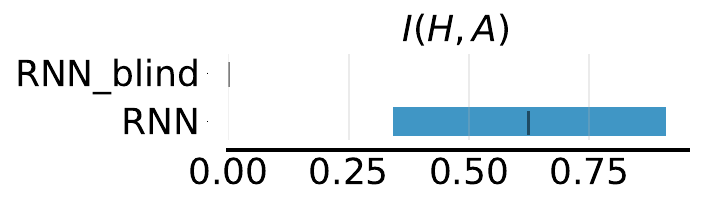}
        \caption{\(I(H, A)\)}
        \label{fig:two_heuristic_agents_sub_duplicate}
    \end{subfigure}

    \caption{Partially Concurrent Learning Agents Experiment: (a) Learning performances; (b) Mutual Information between actions and observations; (c) Mutual Information between agents' actions and histories. We report the mean and bootstrapped $95\%$ Confidence Intervals (CI) over 10 seeds.}
    \label{fig:three_side_by_side}
\end{figure}

\paragraph*{On the Necessity of Grounded Policies.}~~Finally, we investigate whether agents can learn robust policies when the task requires it. To do so, we train agents from scratch in the environment with two fixed partners, deterministically selecting different actions in a repetitive way by following specific cycles (see Appendix~\ref{app:heuristic_agents} for details). Since these agents start each episode in a random phase of their cycle, a learning agent cannot rely on a pre-arranged convention and is forced to infer the hidden state (the current phase and phase length) of its partners from observations. Interestingly, RNN policies do learn effective policies, demonstrating a successful use of memorization to identify the cycles and predict future actions %
In contrast, memoryless MLP policies perform worse, and blind agents fail to model the non-learning agents (Fig.~\ref{fig:four_learning_agents_sub}). Importantly, in both cases the MI between observations and actions (Fig.~\ref{fig:two_heuristic_agents_sub}) and the hidden state and actions (Fig.~\ref{fig:two_heuristic_agents_sub_duplicate}) is now significantly higher. This indicates that, unlike in the previous setting, the agent's policy is now actively and necessarily grounded in its observation history: the mechanism for success has changed.

\section{Do Modern Environments Require Complex Reasoning Relying on Markovian Information?}\label{sec:reasoning}

Section~\ref{sec:case_study} revealed a crucial principle: When an environment is structured to prevent convention-based shortcuts, agents are capable of learning the desired, more complex behaviours of extracting Markovian signals from the environment. This naturally leads to a critical question for the broader MARL community:

\begin{tcolorbox}[colback=softbluegray, colframe=softbluegray,  boxrule=0.5pt, arc=4pt, width=\linewidth]
\begin{center}
\emph{Do modern MARL environments \textbf{actually} require (1) behaviours grounded in observations and (2) memory-based reasoning about other agents?}
\end{center}
\end{tcolorbox}

The answer to this question is paramount: Demanding behaviour that is grounded in observation is necessary to prevent agents from learning brittle conventions, while requiring memory-based reasoning ensures a task truly embodies the challenges of a Dec-POMDP. Without environments that enforce both, we risk measuring progress on benchmarks that can be solved with the same non-generalisable shortcuts identified in our case study. In this section, we investigate a small sample of popular MARL benchmarks to explore this question.

\paragraph*{Agent Modeling: Hanabi.}~~As a first environment instance, we consider Hanabi~\citep{bard2020hanabi}\footnote{In the following, we use the two-player version of this game, where the maximum score is 25.}. It is a partially observable cooperative MARL environment based on the card game, where players see their teammates' cards but not their own. To succeed, players need to exchange clues and use them to infer information about the cards in their possession. The core challenge moves beyond a clue's literal meaning to inferring the teammate's intent---that is, why that specific clue was given. This has made Hanabi a popular benchmark when testing theory of mind, agents' modelling or ad-hoc coordination~\citep{hu2020other,foerster2019bayesian,nekoei2023towards,hu2021off}.

\begin{figure}[t]
  \centering
  \begin{subfigure}[b]{0.32\textwidth}
    \centering
    \includegraphics[height=3cm]{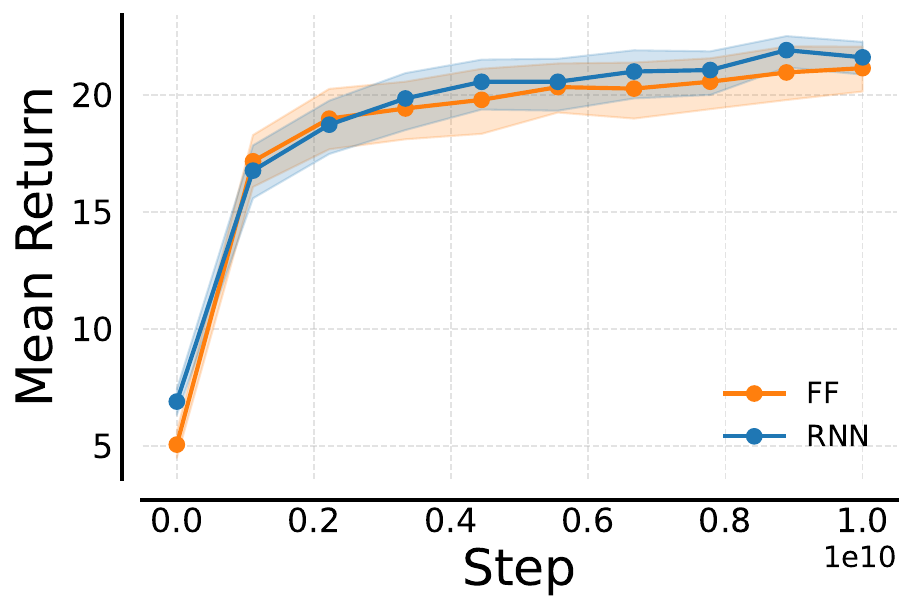}
    \caption{Hanabi (Two-Players).}
    \label{fig:hanabi-mean}
  \end{subfigure}\hfill
  \begin{subfigure}[b]{0.32\textwidth}
    \centering
    \includegraphics[height=3cm]{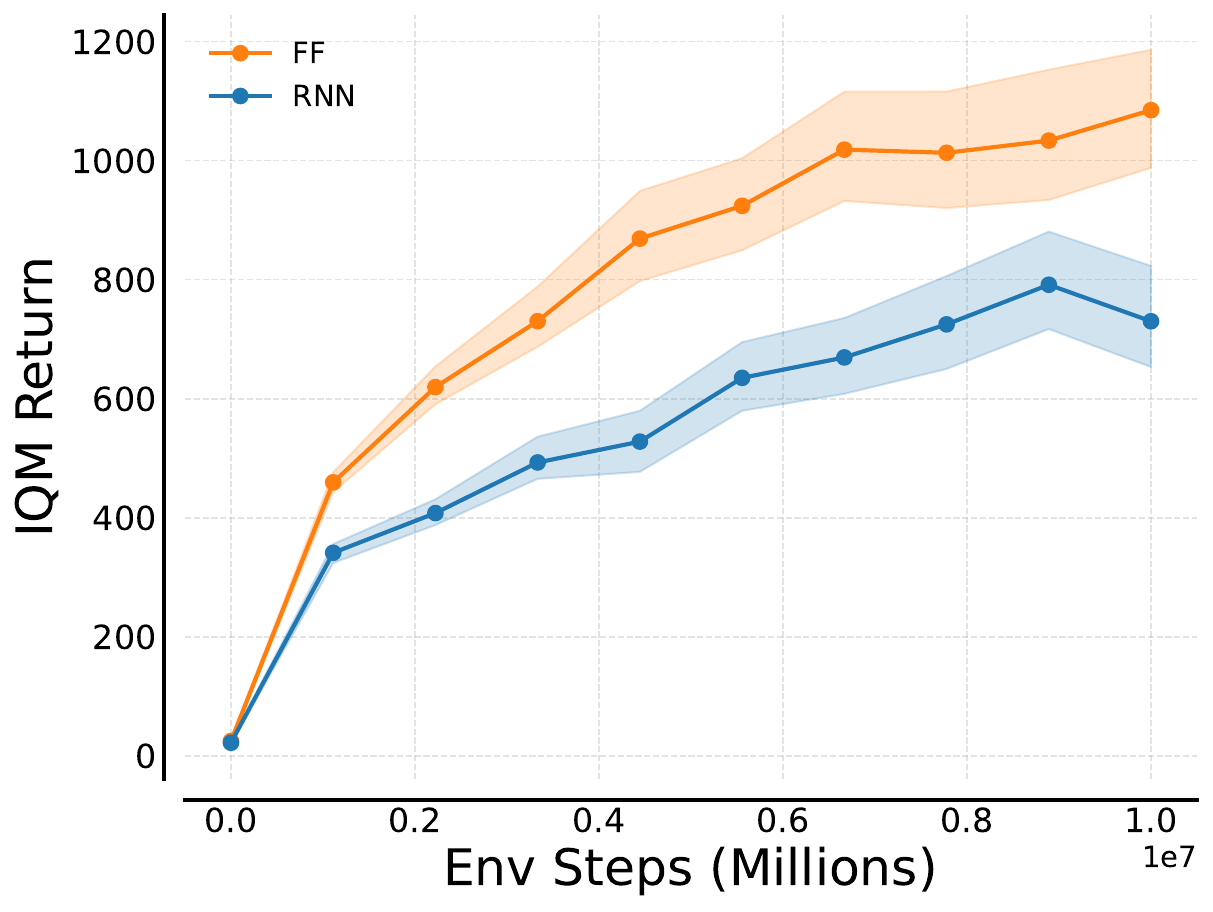}
    \caption{MaBrax. }
    \label{fig:mabrax-iqm}
  \end{subfigure}\hfill
  \begin{subfigure}[b]{0.32\textwidth}
    \centering
    \includegraphics[height=3cm]{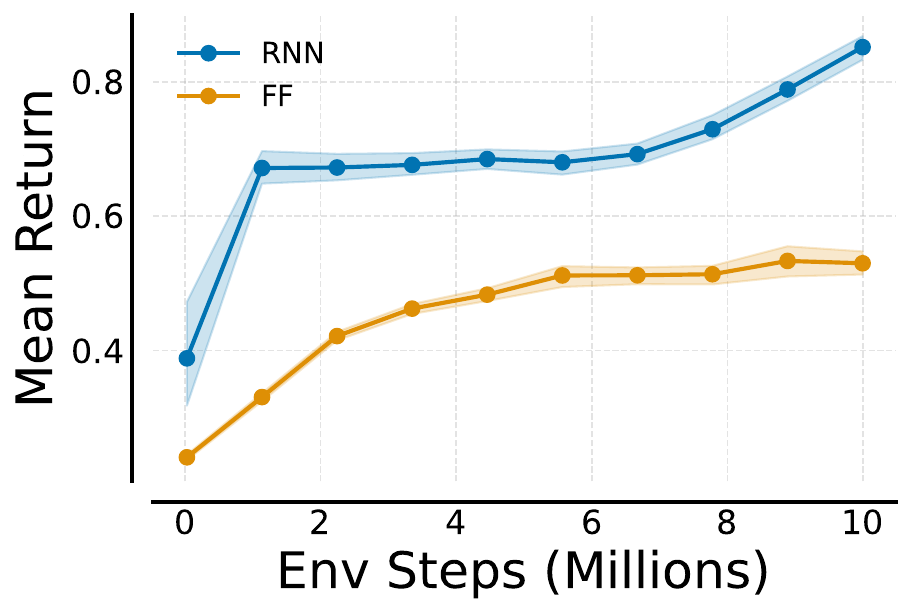}
    \caption{SMAX (5 Units).}
    \label{fig:smax-mean}
  \end{subfigure}

  \vspace{1em}

  \begin{minipage}[t]{0.45\textwidth}
    \centering
    \begin{subfigure}[b]{\textwidth}
      \centering
      \includegraphics[width=0.65\linewidth]{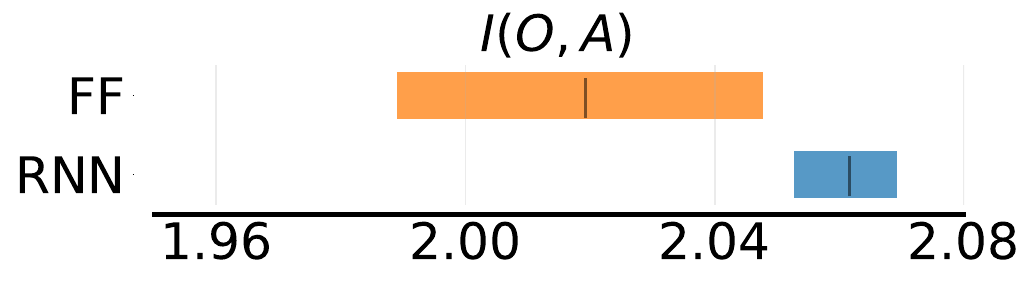}
      \caption{Hanabi ($\mathbb{I}(O;A)_{max}\approx 3$)}
      \label{fig:mi-obs-hanabi}
    \end{subfigure}

    \vspace{0.5em}

    \begin{subfigure}[b]{\textwidth}
      \centering
      \includegraphics[width=0.65\linewidth]{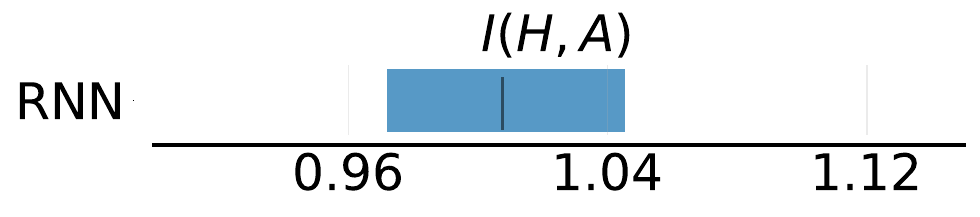}
      \caption{Hanabi ($\mathbb{I}(H;A)_{max}\approx 3$)}
      \label{fig:mi-hid-hanabi}
    \end{subfigure}
  \end{minipage}\hfill
  \begin{minipage}[t]{0.45\textwidth}
    \centering
    \begin{subfigure}[b]{\textwidth}
      \centering
      \includegraphics[width=0.65\linewidth]{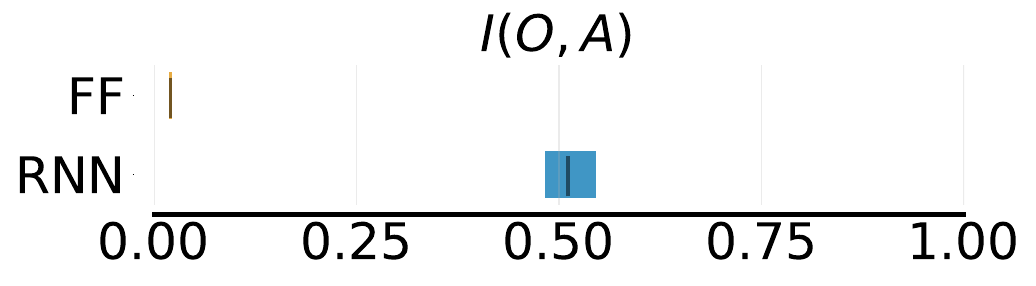}
      \caption{SMAX ($\mathbb{I}(O;A)_{max}\approx 2.30$)}
      \label{fig:mi-obs-smax}
    \end{subfigure}

    \vspace{0.5em}

    \begin{subfigure}[b]{\textwidth}
      \centering
      \includegraphics[width=0.65\linewidth]{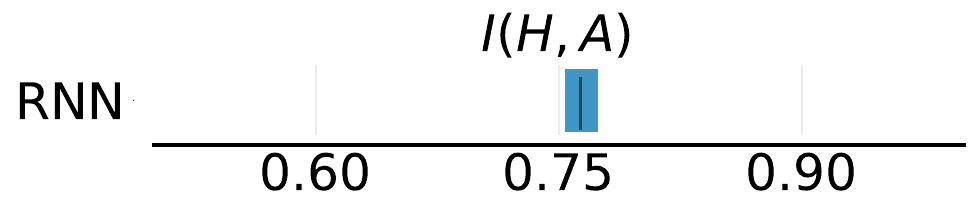}
      \caption{SMAX ($\mathbb{I}(H;A)_{max}\approx 2.30$) }
      \label{fig:mi-hid-smax}
    \end{subfigure}
  \end{minipage}

  \caption{%
    (a–c) Sample-efficiency (interquartile-mean) across Hanabi, MaBrax, and SMAX.  
    (d–g) Mutual information between observations and actions $\!\mathbb{I}(O;A)\!$ and between hidden state and actions $\!\mathbb{I}(H;A)$, stacked per environment for Hanabi (d–e) and SMAX (f–g).
  }
  \label{fig:all-results}
\end{figure}

Since an agent's only source of information comes from the clues provided by its partner, this design encourages policies to be grounded in observation. Our findings confirm this: the MI analysis shows that both RNN and FF policies learn to actively use their observations (Fig.~\ref{fig:mi-obs-hanabi}) or the RNN hidden state (Fig.~\ref{fig:mi-hid-hanabi}). However, our results also show that FF achieves nearly identical performance to the RNN, suggesting that memory provides no significant advantage in this specific task instantiation (see Fig.~\ref{fig:hanabi-mean}). This implies that while Hanabi successfully necessitates grounded policies, it does not adequately test for complex, memory-based reasoning about other agents over time. The task can be solved with no notion of history, revealing a limitation in its ability to evaluate the full spectrum of reasoning required in complex Dec-POMDPs.\footnote{As a side note, we highlight that throughout the experiments agents' observations included the entire discard pile and not just the top card. Different observation functions might call for the need for memorization.}\vspace{-0.3cm}

\paragraph*{Continuous Control: MABrax.}~~To explore our central questions in more complex, continuous control MARL, we next evaluate our policies on five difference instances of Multi-Agent Brax~\citep[MABrax,][]{rutherford2023jaxmarl}. MABrax is a JAX-accelerated version of the MaMuJoCo~\citep {peng2021facmac}, where a robot's body parts are controlled by different agents. Each agent receives only ego-centric observations, such as its own joint angles and velocities, along with those of its immediate neighbours, with the goal being to collaborate to move forward.

The results reported in Figure~\ref{fig:mabrax-iqm} present a potentially surprising result:\footnote{The interested reader can refer to Figure~\ref{fig:performance_mabrax_detailed} in Appendix for the detailed plots.} memoryless feed-forward architectures outperform their RNN counterparts. We posit that this is because in locomotion-style tasks, current proprioception fully determines the optimal torque, making memory less relevant for these tasks. Furthermore, limited observability of distant agents also reduces any incentive for complex partner modelling. As a side note, we highlight that while MaMuJoCo-inspired environments are currently ubiquitous as multi-agent benchmarking, they hide a crucial limitation: agents completely unaware of others can \emph{still} achieve non-trivial performances in a set of instances. We reported these results in Fig.~\ref{fig:blind_mabrax_detailed} in the Appendix, unlike previous results by~\citet{ellis2023smacv2}, we also removed remove time steps from agents' observations. 

Overall, these results provide strong evidence that popular locomotion benchmarks may not adequately evaluate challenging multi-agent learning. We also caution against using the fully-observable variants of these environments found in some recent work~\citep{wangorder,zhong2024heterogeneous}, as they deviate from the Dec-POMDP problem structure.\vspace{-0.3cm}

\paragraph*{Meaningful Partial Observability: SMAX.}~~As a final experiment, we evaluate IPPO agents on SMAX~\citep{rutherford2023jaxmarl}, a JAX-accelerated version of the SMAC~\citep{samvelyan2019starcraft}. Specifically, we focus on SMAC-v2~\citep{ellis2023smacv2}, which was introduced to address limitations in the original SMAC, such as a simple open-loop policies could succeed while ignoring observations. In contrast, SMAC-v2 incorporates stochastic starting positions and enforcing "meaningful partial observability", where agents must infer critical information held by their teammates.

Our results confirm that these changes successfully create a memory-dependent task. As shown in Figure~\ref{fig:smax-mean}, RNN policies outperform their FF counterparts by a large margin, a finding also noted by~\cite{ellis2023smacv2}. However, a deeper analysis reveals a more nuanced picture. When we look at $\mathbb{I}(O;A)$ and $\mathbb{I}(H;A)$, we see a moderate dependence between these variables and actions, which correspond to approximately 22\% and 33\% of the theoretical maximum $H(A) = \ln 10 \approx 2.30$ for a 10-action space, which is significantly lower than in coordination-centric benchmarks like Hanabi (69\% and 32\% of their maximum). This suggests that while SMAC-v2 maps are a significant improvement, there is still potential to design environments that demand an even stronger reliance on history-based reasoning to fully capture the complexity of Dec-POMDPs.

\section{Conclusions and Takeaway}

In this paper, we provide empirical evidence that suggests robust and generalisable MARL systems are fundamentally gated by the design of evaluation environments: These environments must be designed to necessitate and reward policies grounded in the known hardness of Dec-POMDPs (Fig.~\ref{fig:subfigureExample}). By removing the possibility of convention-based shortcuts, we showed in Section~\ref{sec:case_study} that agents can be encouraged to develop more meaningful and grounded policies. Additionally, we showed in Section~\ref{sec:reasoning}, that many common benchmarks do not require temporal reasoning or grounded policies.

\textbf{Takeaway.}~~We advocate for the design and adoption of benchmarks that compel agents to develop policies built upon two core principles: (1) \textbf{\textcolor{softbluegray3}{behaviours grounded in observations}} and (2) \textbf{\textcolor{softbluegray3}{memory-based reasoning}} about other agents. We claim that enforcing these properties is essential to ensure that environments capture the true complexities of Dec-POMDPs and drive the development of more reliable multi-agent systems. While modern benchmarks show promise, our analysis indicates they too can be improved to more rigorously test these foundational capabilities.

\textbf{Next-steps.}~~We acknowledge several avenues for future research. Our analysis focused on Independent PPO (IPPO) with non-shared parameters, and an important next step would be to investigate how these findings extend to algorithms with parameter sharing and methods with centralised critics such as MAPPO~\citep{yu2022surprising}. Furthermore, while our study spanned several distinct environments, extending this diagnostic approach across an even wider range of MARL benchmarks would help to build a more comprehensive understanding of the MARL evaluation landscape.

\section{Acknowledgements}
We would like to thank Aris Filos-Ratsikas for fruitful discussions on early versions of this work. An author on this project has received funding towards this work from the European Union’s Horizon Europe research and innovation programme under grant agreement No. 101120726. 
\bibliography{main}
\bibliographystyle{rlj}

\newpage
\appendix
\section{Appendix}
\subsection{Prediction Game: Heuristic Agents}
\label{app:heuristic_agents}

A heuristic (non-learning) agent $i$ follows a simple periodic policy. At environment step $t$ its action is
\[
a^i(t)
  =\Bigl(\,(i \bmod A)\;+\;
         \Bigl\lfloor \frac{t}{k_i}\Bigr\rfloor
         +\phi_i\Bigr)\bmod A ,
\]
where  
\begin{itemize}
  \item $A = |\mathcal{A}|$ is the number of discrete actions;
  \item $k_i \in \mathbb{N}{>0}$ is the \textbf{cycle length}: the agent repeats the same action for $k_i$ steps before advancing to the next in a modulo-$A$ loop;
  \item $\phi_i \sim \mathrm{Uniform}\{0,\dots,k_i-1\}$ is a fresh \textbf{initial phase} drawn at the start of every episode, so each episode begins at a random point in the cycle.
\end{itemize}

\paragraph{Example.}
Assume $A=4$ (actions $\{0,1,2,3\}$).  
\begin{itemize}
  \item \textbf{Agent 0} with $k_0=3$ and $\phi_0=1$ executes the sequence  
        \[
          (1,1,1,\;2,2,2,\;3,3,3,\;0,0,0,\dots).
        \]
  \item \textbf{Agent 2} with $k_2=2$ and $\phi_2=0$ executes  
        \[
          (2,2,\;3,3,\;0,0,\;1,1,\;2,2,\dots).
        \]
\end{itemize}

This produces predictable but non-trivial periodic behaviour. Furthermore, learning agents only have $1/4$ of selecting the first action correctly since there is no previous action to help them make this selection and no information from previous episodes that can be used to determine the starting point in a cycle. 

\subsection{Prediction Game: Additional Plots}
\begin{figure}[htbp]
  \centering
  \begin{subfigure}[b]{0.75\textwidth}
    \centering
    \includegraphics[width=\linewidth]{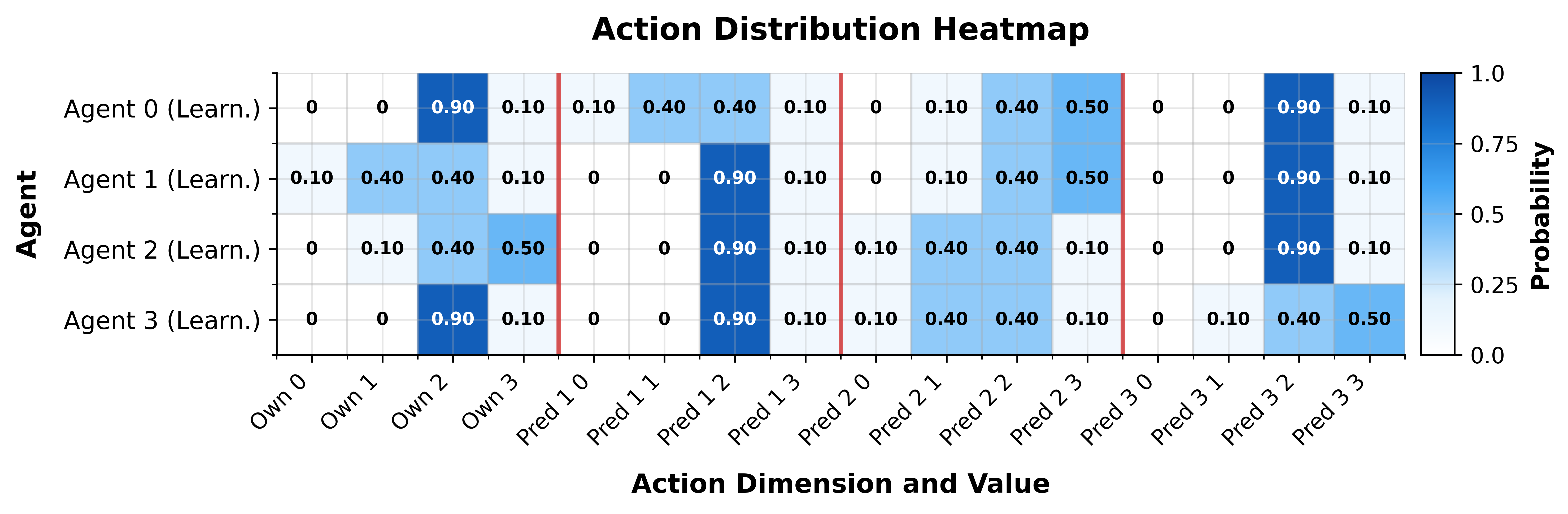}
    \caption{FF Action Histogram}
    \label{fig:image1}
  \end{subfigure}
  
  \vspace{1em} %

  \begin{subfigure}[b]{0.75\textwidth}
    \centering
    \includegraphics[width=\linewidth]{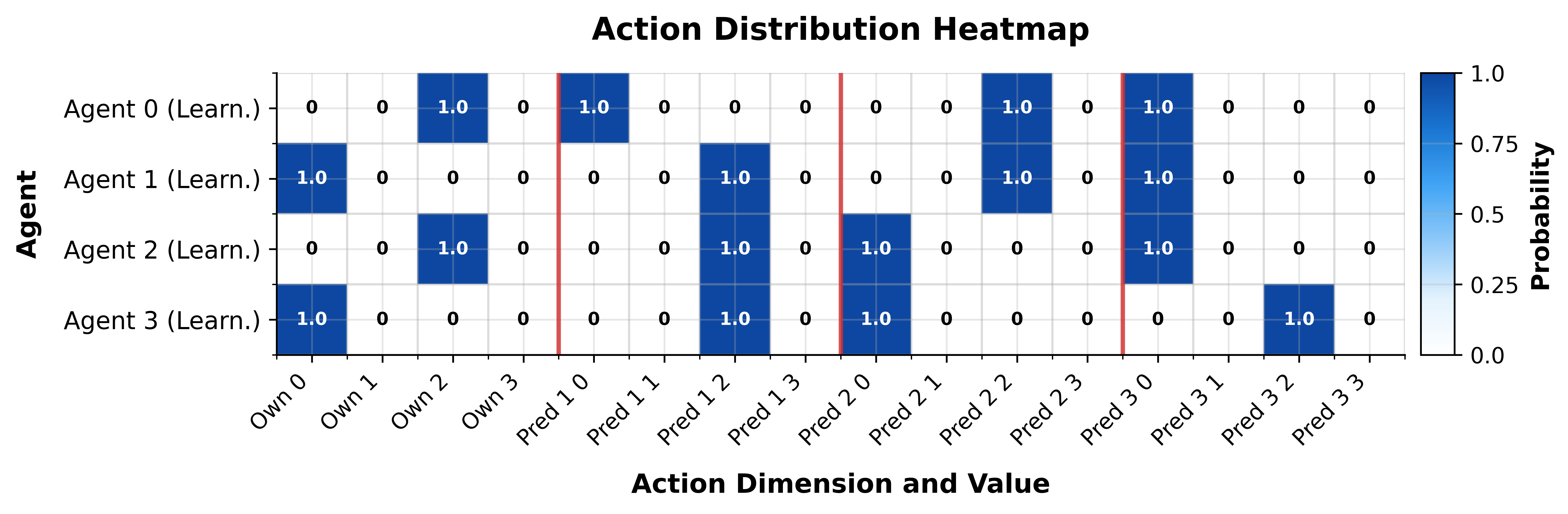}
    \caption{RNN Action Histogram}
    \label{fig:image2}
  \end{subfigure}
  
  \caption{Action Histograms for FF and RNN IPPO agents on the Prediction Game, evaluated over 1000 episodes at the end of training.}
  \label{fig:action_dist_pred_game}
\end{figure}

In Figure~\ref{fig:action_dist_pred_game}, we show the action histograms of our concurrently learning agents during evaluation. We see that they form conventions and stick to specific action patterns.

\subsection{MaBrax: Additional Results}\label{app:mabrax_additional}

\begin{figure*}[h!]
    \centering
    \begin{subfigure}[b]{0.3\textwidth}
        \centering
        \includegraphics[height=3.5cm]{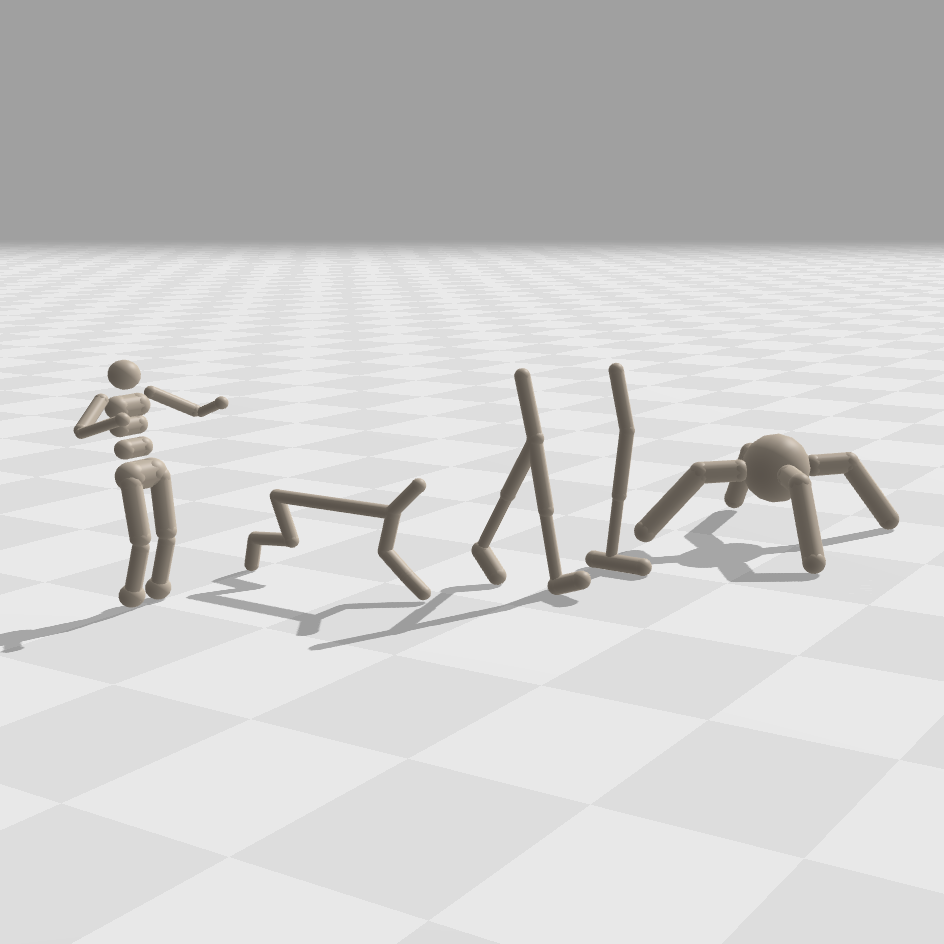}
        \caption{MABrax environments~\citep{peng2021facmac,rutherford2023jaxmarl}}
        \label{fig:mabrax_env}
    \end{subfigure}
    \hfill
    \begin{subfigure}[b]{0.3\textwidth}
        \centering
        \includegraphics[width=\textwidth]{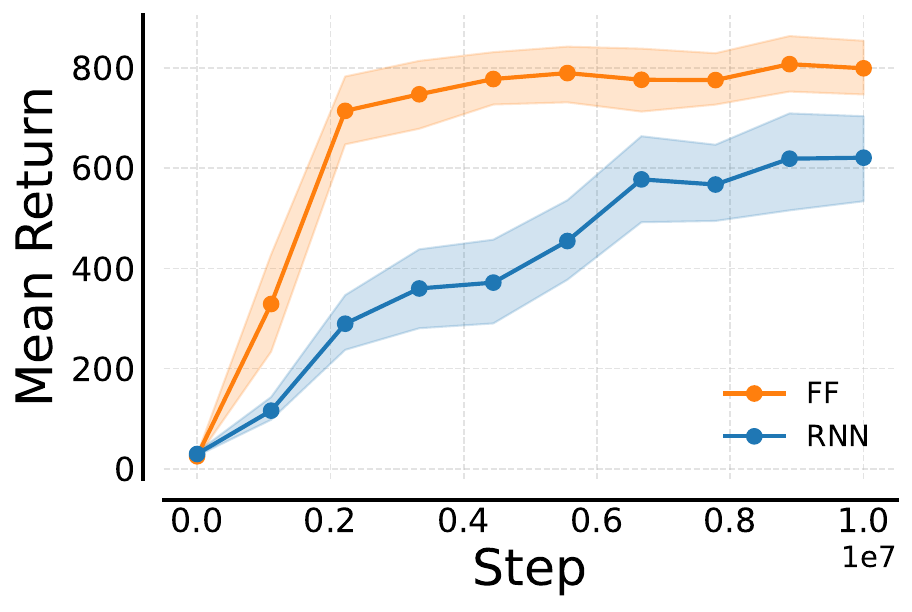}
        \caption{Ant $4\times2$}
        \label{fig:mabrax_1}
    \end{subfigure}
    \hfill
    \begin{subfigure}[b]{0.3\textwidth}
        \centering
        \includegraphics[width=\textwidth]{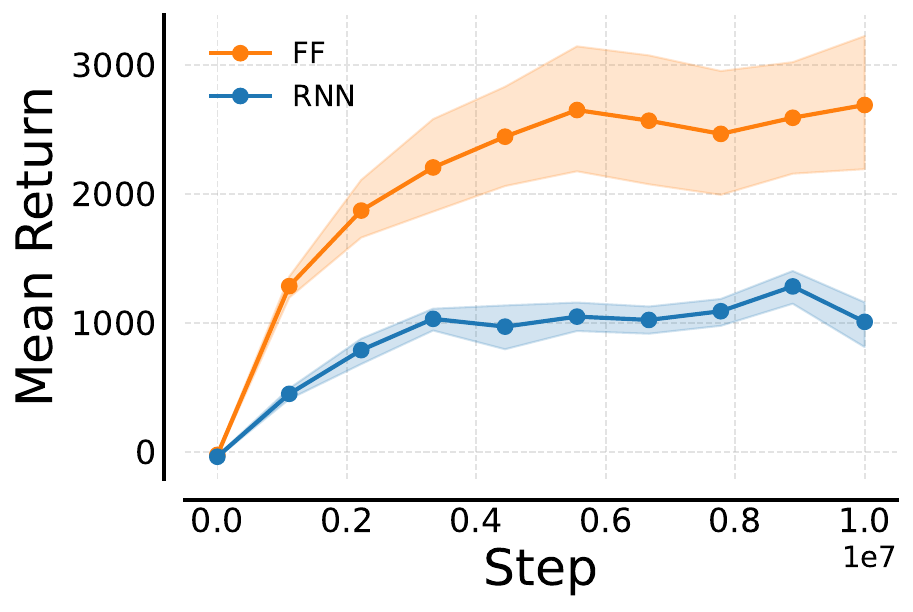}
        \caption{Half Cheetah $6\times1$}
        \label{fig:mabrax_2}
    \end{subfigure}

    \vspace{1.5em}

    \begin{subfigure}[b]{0.3\textwidth}
        \centering
        \includegraphics[width=\textwidth]{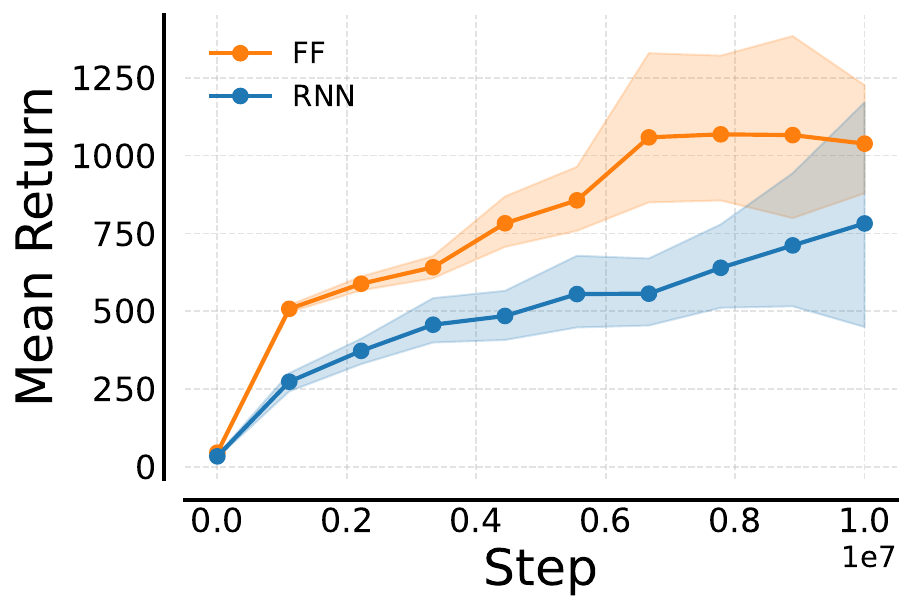}
        \caption{Hopper $3 \times 1$}
        \label{fig:mabrax_3}
    \end{subfigure}
    \hfill
    \begin{subfigure}[b]{0.3\textwidth}
        \centering
        \includegraphics[width=\textwidth]{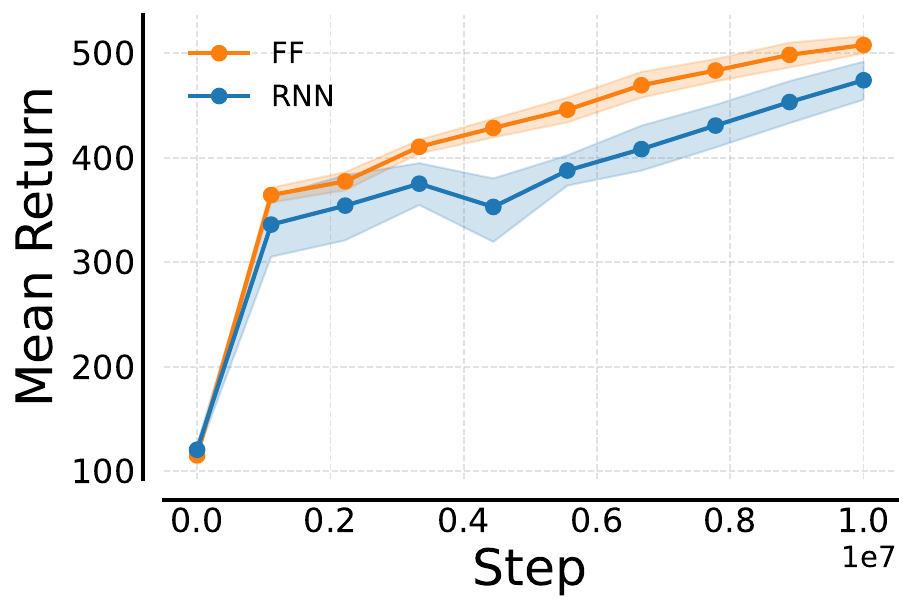}
        \caption{Humanoid 8|9}
        \label{fig:mabrax_4}
    \end{subfigure}
    \hfill
    \begin{subfigure}[b]{0.3\textwidth}
        \centering
        \includegraphics[width=\textwidth]{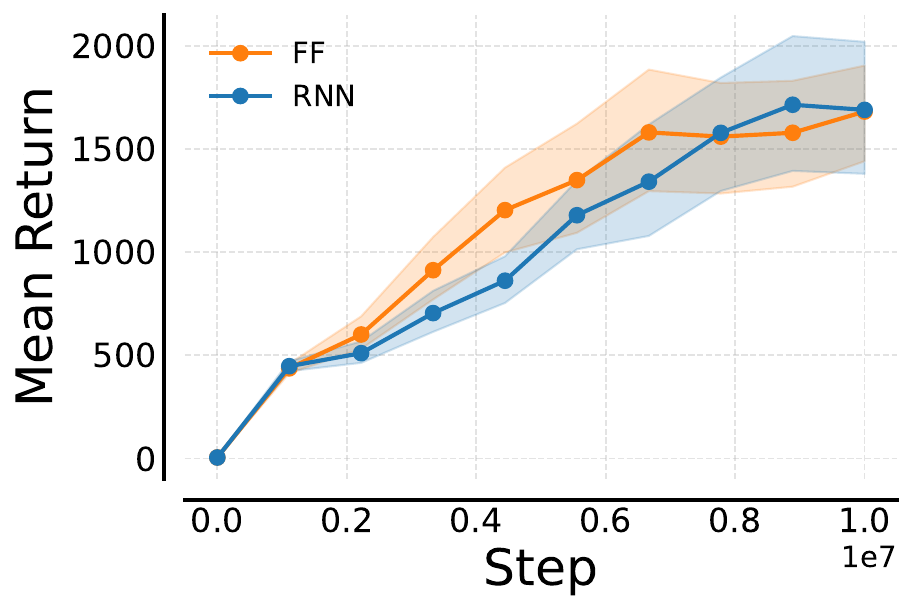}
        \caption{Walker $2\times3$}
        \label{fig:mabrax_5}
    \end{subfigure}

    \caption{Top: MABrax environment suite and sample tasks. Bottom: means and 95\% bootstrapped confidence intervals across 16 seeds for each environment.}
    \label{fig:performance_mabrax_detailed}
\end{figure*}

\begin{figure*}[h]
    \centering
    \begin{subfigure}[b]{0.3\textwidth}
        \centering
        \includegraphics[width=\textwidth]{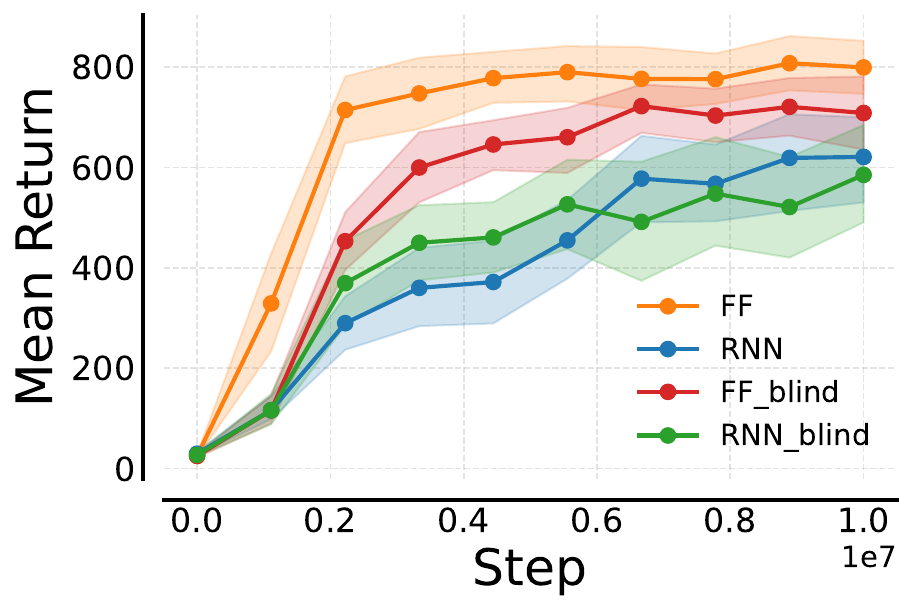}
        \caption{Ant $4\times2$}
        \label{fig:gobs_mabrax_1}
    \end{subfigure}
    \hfill
    \begin{subfigure}[b]{0.3\textwidth}
        \centering
        \includegraphics[width=\textwidth]{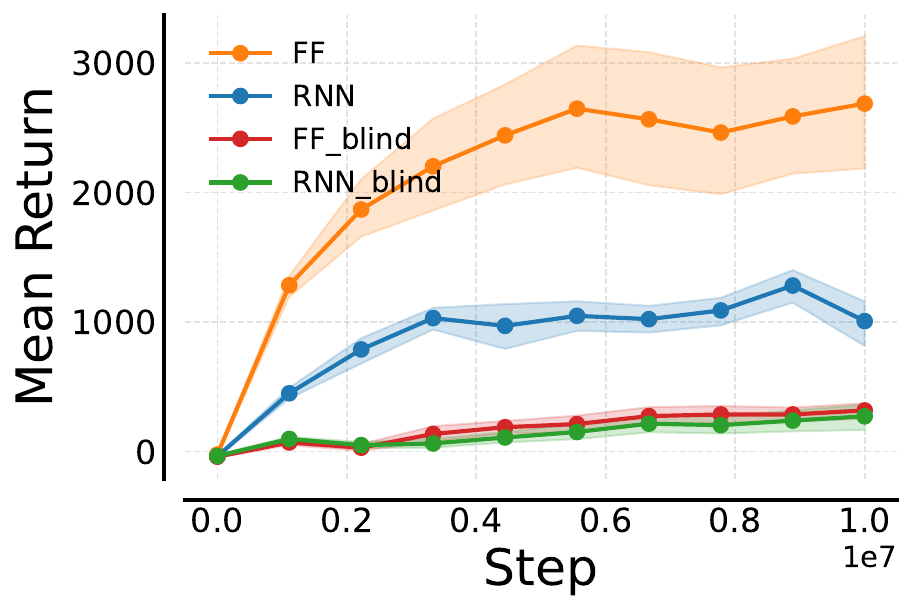}
        \caption{Half Cheetah $6\times1$}
        \label{fig:all_mabrax_2}
    \end{subfigure}
    \hfill
    \begin{subfigure}[b]{0.3\textwidth}
        \centering
        \includegraphics[width=\textwidth]{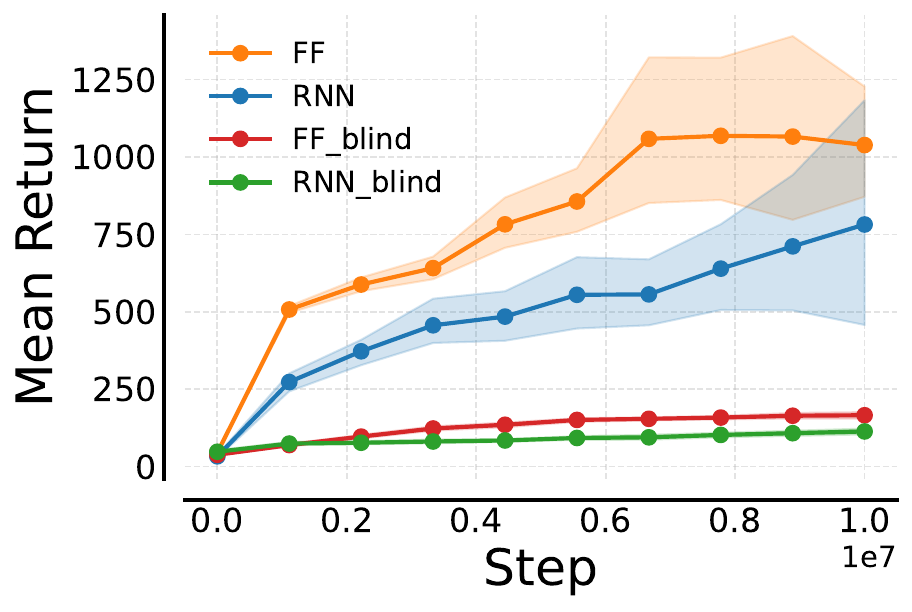}
        \caption{Hopper $3 \times 1$}
        \label{fig:all_mabrax_3}
    \end{subfigure} \\
    \vfill
    \begin{subfigure}[b]{0.3\textwidth}
        \centering
        \includegraphics[width=\textwidth]{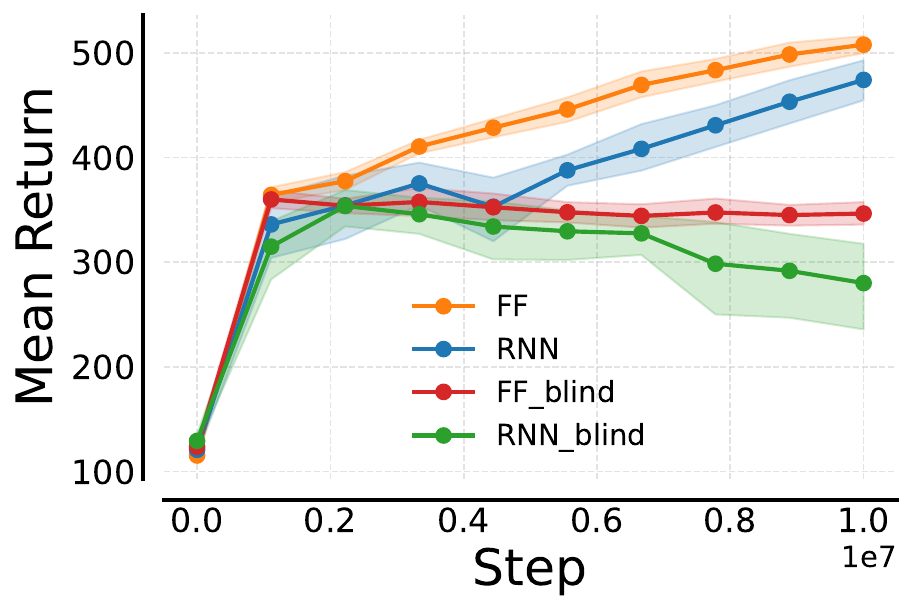}
        \caption{Humanoid 8|9}
        \label{fig:all_mabrax_4}
    \end{subfigure}
        \begin{subfigure}[b]{0.3\textwidth}
        \centering
        \includegraphics[width=\textwidth]{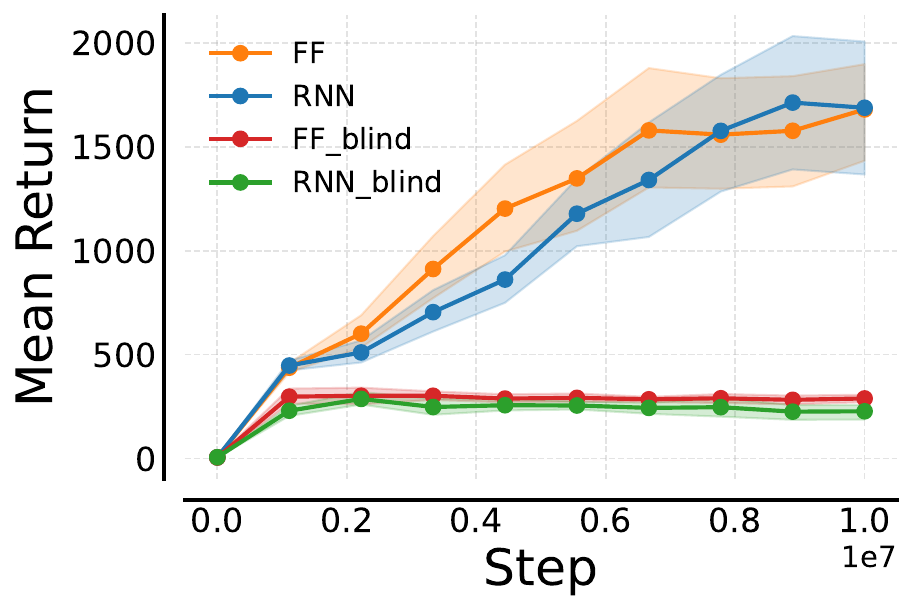}
        \caption{Walker $2\times3$}
        \label{fig:all_mabrax_5}
    \end{subfigure}
    \caption{MABrax environments comparing partially-observable, to blind performances showcasing mean returns and 95\% bootstrapped confidence intervals across 16 seeds}
    
    \label{fig:blind_mabrax_detailed}
\end{figure*}

Here we show more detailed results of the individual MaBrax tasks in figure \ref{fig:performance_mabrax_detailed}. In \ref{fig:blind_mabrax_detailed} we show a comparison of results of blind agents who don't receive an observation at all.

\subsection{Hyperparameters}

\subsubsection{Prediction Game}

For each experiment in the Prediction Game, we perform a sweep over the following hyperparameters: learning rate ($\text{LR} \in \{1\times10^{-4},\, 3\times10^{-4},\, 5\times10^{-4},\, 1\times10^{-3}\}$), clipping epsilon ($\text{CLIP\_EPS} \in \{0.1,\, 0.2,\, 0.5\}$), and whether to anneal the learning rate ($\text{ANNEAL\_LR} \in \{\text{True},\, \text{False}\}$). We report the best performance per method in Table~\ref{tab:pred_game}.

\begin{table}[htbp]
\centering
\caption{Hyperparameters used for Prediction Game}
\label{tab:pred_game}
\begin{tabular}{lcc}
\hline
\textbf{Hyperparameter} & \textbf{RNN} & \textbf{FF} \\
\hline
\multicolumn{3}{l}{\textbf{Network parameters}} \\
Agent parameter sharing & False & False \\
Embedding dimension & 128 & - \\
GRU hidden dimension & 128 & - \\
Actor hidden dimension & - & 128 \\
Critic hidden dimension & - & 128 \\
Activation function & relu & relu \\
\hline
\multicolumn{3}{l}{\textbf{Training parameters}} \\
Total time steps & $1.0 \times 10^{7}$ & $1.0 \times 10^{7}$ \\
Number of steps & 128 & 128 \\
Number of environments & 16 & 16 \\
Number of evaluation episodes & - & - \\
Number of seeds & 10 & 10 \\
Update epochs & 4 & 4 \\
Number of minibatches & 4 & 4 \\
Learning rate annealing & False & True \\
Learning rate & $5.0 \times 10^{-4}$ & $2.5 \times 10^{-4}$ \\
Entropy coefficient & $1.0 \times 10^{-2}$ & $1.0 \times 10^{-2}$ \\
Clipping epsilon & 0.2 & 0.2 \\
Scale clipping epsilon & False & - \\
Ratio clipping epsilon & - & - \\
Gamma & 0.99 & 0.99 \\
GAE lambda & 0.95 & 0.95 \\
Value function coefficient & 0.5 & 0.5 \\
Max gradient norm & 0.5 & 0.5 \\
\hline
\end{tabular}
\end{table}

\subsubsection{MaBrax}

For each MaBrax environment, we perform random sweeps over 32 different learning rates, where $\text{LR} \in [0.0001, 0.01]$. Each selected learning rate is evaluated across 5 random seeds, separately for both the RNN and feed-forward (FF) implementations. The final selected hyperparameters are provided in Table~\ref{tab:mabrax_hyperparameters}.

\begin{table}[htbp]
\centering
\caption{Hyperparameters used for MaBrax Experiments}
\label{tab:mabrax_hyperparameters}
\begin{tabular}{lcc}
\hline
\textbf{Hyperparameter} & \textbf{RNN} & \textbf{FF} \\
\hline
\multicolumn{3}{l}{\textbf{Network parameters}} \\
Agent parameter sharing & False & False\\
Embedding dimension & 128 & - \\
GRU hidden dimension & 128 & - \\
Actor hidden dimension & - & 128 \\
Critic hidden dimension & - & 128 \\
Activation function & tanh & tanh \\
\hline
\multicolumn{3}{l}{\textbf{Training parameters}} \\
Total time steps & $1.0 \times 10^{7}$ & $1.0 \times 10^{7}$ \\
Number of steps & 64 & 64 \\
Number of environments & 256 & 256\\
Number of evaluation episodes & 32 & 32 \\
Number of seeds & 16 & 16 \\
Update epochs & 4 & 4 \\
Number of minibatches & 4 & 4 \\
Learning rate annealing & False & False \\
Learning rate Ant $2 \times 4$ & $3.0 \times 10^{-3}$ & $3.0 \times 10^{-4}$ \\
Learning rate Half-Cheetah $6 \times 1$ & $2.1 \times 10^{-4}$ & $1.0 \times 10^{-3}$ \\
Learning rate Hopper $3 \times 1$ & $1.0 \times 10^{-3}$ & $2.5 \times 10^{-3}$ \\
Learning rate Humanoid 8|9 & $1.8 \times 10^{-3}$ & $8.5 \times 10^{-4}$ \\
Learning rate Walker2d $2 \times 3$ & $9.5 \times 10^{-4}$ & $3.4 \times 10^{-3}$ \\
Entropy coefficient & $1.0 \times 10^{-4}$ & $1.0 \times 10^{-4}$ \\
Clipping epsilon & 0.2 & 0.2 \\
Scale clipping epsilon & False & False \\
Ratio clipping epsilon & False & False \\
Gamma & 0.99 & 0.99 \\
GAE lambda & 0.95 & 0.95 \\
Value function coefficient & 1.0 & 1.5 \\
Max gradient norm & 0.5 & 0.5 \\
Adam epsilon & $1.0 \times 10^{-8}$ & $1.0 \times 10^{-8}$ \\
Advantage unroll depth & 8 & 8 \\
\hline
\end{tabular}
\end{table}

\subsubsection{Hanabi}

For Hanabi we use the same hyperparameter settings as JaxMARL \citep{rutherford2023jaxmarl}, these can be found in Table \ref{tab:hanabi_hyperparameters} below. 

\begin{table}[htbp]
\centering
\caption{Hyperparameters used for Hanabi Experiments}
\label{tab:hanabi_hyperparameters}
\begin{tabular}{lcc}
\hline
\textbf{Hyperparameter} & \textbf{FF} & \textbf{RNN} \\
\hline
\multicolumn{3}{l}{\textbf{Network parameters}} \\
Agent parameter sharing & False & False \\
Embedding dimension & - & 128 \\
GRU hidden dimension & - & 128 \\
Actor hidden dimension & 128 & - \\
Critic hidden dimension & 128 & - \\
Activation function & tanh & tanh \\
\hline
\multicolumn{3}{l}{\textbf{Training parameters}} \\
Total time steps & $1.0 \times 10^{10}$ & $1.0 \times 10^{10}$ \\
Number of steps & 128 & 128 \\
Number of environments & 1024 & 1024 \\
Number of evaluation episodes & 128 & 128 \\
Number of seeds & 8 & 8 \\
Number of checkpoints & 256 & 256 \\
Update epochs & 4 & 4 \\
Number of minibatches & 4 & 4 \\
Learning rate annealing & True & True \\
Learning rate & $5.0 \times 10^{-4}$ & $5.0 \times 10^{-4}$ \\
Entropy coefficient & $1.0 \times 10^{-2}$ & $1.0 \times 10^{-2}$ \\
Clipping epsilon & 0.2 & 0.2 \\
Scale clipping epsilon & False & False \\
Ratio clipping epsilon & False & False \\
Gamma & 0.99 & 0.99 \\
GAE lambda & 0.95 & 0.95 \\
Value function coefficient & 1.0 & 1.0 \\
Max gradient norm & 0.5 & 0.5 \\
Adam epsilon & $1.0 \times 10^{-8}$ & $1.0 \times 10^{-8}$ \\
Advantage unroll depth & 8 & 8 \\
\hline
\end{tabular}
\end{table}

\subsubsection{SMAX}

For SMAX we use the hyperparameters from JAxMARL \cite{rutherford2023jaxmarl}, as shown in Table \ref{tab:smax_hyperparameters}.

\begin{table}[htbp]
\centering
\caption{Hyperparameters used for SMAX Experiments}
\label{tab:smax_hyperparameters}
\begin{tabular}{lcc}
\hline
\textbf{Hyperparameter} & \textbf{FF} & \textbf{RNN} \\
\hline
\multicolumn{3}{l}{\textbf{Network parameters}} \\
Recurrent & False & True \\
GRU hidden dimension & - & 128 \\
Fully connected dimension & - & 128 \\
Activation function & relu & relu \\
\hline
\multicolumn{3}{l}{\textbf{Training parameters}} \\
Total time steps & $1.0 \times 10^{7}$ & $1.0 \times 10^{7}$ \\
Number of steps & 128 & 128 \\
Number of environments & 128 & 128 \\
Update epochs & 4 & 2 \\
Number of minibatches & 4 & 2 \\
Learning rate annealing & True & True \\
Learning rate & $4.0 \times 10^{-3}$ & $4.0 \times 10^{-3}$ \\
Entropy coefficient & 0.0 & 0.0 \\
Clipping epsilon & 0.1 & 0.2 \\
Scale clipping epsilon & - & False \\
Gamma & 0.99 & 0.99 \\
GAE lambda & 0.95 & 0.95 \\
Value function coefficient & 0.5 & 0.5 \\
Max gradient norm & 0.5 & 0.5 \\
Seed & 30 & 30 \\
Map name & smacv2\_5\_units & smacv2\_5\_units \\
See enemy actions & True & True \\
Walls cause death & True & True \\
Attack mode & closest & closest \\
Max steps & 100 & 100 \\
\hline
\end{tabular}
\end{table}

\end{document}